\documentclass[preprint,12pt]{elsarticle}

\usepackage{booktabs}
\usepackage{caption}
\usepackage{graphicx}
\graphicspath{{Figures_new/}}
\usepackage{amsmath}
\usepackage{amssymb}
\usepackage{subfigure}
\usepackage[marginal]{footmisc}

\usepackage{hyperref}

\journal{Information sciences}

\begin{document}

\begin{frontmatter}



\author[label1]{Shiming chen}
\ead{gchenshiming@gmail.com}
\author[label1,label2]{Yisong Wang\corref{cor1}}
\ead{yswang@gzu.edu.cn}
\author[label3]{Chin-Teng Lin}
\author[label3,label4]{Weiping Ding}
\author[label3]{Zehong Cao\corref{cor1}}
\address[label1]{School of Computer Science and Technology, Guizhou University, Guizhou, China}
\address[label2]{Key Laborary of Intelligent Medical Image Analysis and Precise Diagnosis of Guizhou Province, Guizhou, China}
\address[label3]{Centre for Artifcial Intelligence, Faculty of Engineering and
	IT, University of Technology Sydney, NSW, Australia}
\address[label4]{School of Computer Science and Technology, Nantong University, Jiangsu, China}

\ead{zehong.cao@uts.edu.au}
\cortext[cor1]{Corresponding authors to Yisong Wang and Zehong Cao.}

\title{Semi-supervised Feature Learning For Improving Writer Identification}


\begin{abstract}
Data augmentation is usually used by supervised learning approaches for offline writer identification, but such approaches require extra training data and potentially lead to overfitting errors. In this study, a semi-supervised feature learning pipeline was proposed to improve the performance of writer identification by training with extra unlabeled data and the original labeled data simultaneously. Specifically, we proposed a weighted label smoothing regularization (WLSR) method for data augmentation, which assigned the weighted uniform label distribution to the extra unlabeled data. The WLSR method could regularize the convolutional neural network (CNN) baseline to allow more discriminative features to be learned to represent the properties of different writing styles. The experimental results on well-known benchmark datasets (ICDAR2013 and CVL) showed that our proposed semi-supervised feature learning approach could significantly improve the baseline measurement and perform competitively with existing writer identification approaches. Our findings provide new insights into offline write identification.

\end{abstract}

\begin{keyword}
Semi-Supervised Learning, Feature Extraction, Regularization, CNN, Writer Identification.

\end{keyword}

\end{frontmatter}


\section{Introduction}
Handwritten texts, speech, fingerprints, and faces are often applied in physiological biometric identifiers. Especially, handwritten text plays an important role for forensics and security in proving someone's authenticity. Research into writer identification has received renewed interest in recent years, such as historical document analysis for the mass-digitization processes of historical documents \cite{Kleber2013CVL,Louloudis2013ICDAR,Xing2016DeepWriter} through machine learning; unfortunately, this process requires considerable time and detection costs. Therefore, many researchers have proposed state-of-the-art pattern recognition approaches to automatically recognize writing styles \cite{Abdi2015A,Christlein2017Writer,Mallikarjunaswamy2013Writer,Plamondona1989Automatic,Yin2009Handwritten}.

Writer identification aims to search and recognize texts written by the same writer in a query database. Writer identification has been investigated on different handwritten scripts, such as English \cite{Bulacu2007Text,Schomaker2004Automatic}, Chinese \cite{He2015Junction,He2008Writer,Wu2014Offline}, Arabic \cite{Abdi2015A}, Indic \cite{Mallikarjunaswamy2013Writer}, Persian \cite{Helli2010A} and Latin scripts \cite{Christlein2017Unsupervised}.
This task generally presents substantial challenges because it requires the documents to be sorted according to high similarity (e.g., the distance of feature vectors). Writer identification can be classified as online writer identification and offline writer identification according to the handwritten document acquisition method. The latter approach can be further categorized into allograph-based and textual-based methods. Textural-based methods compute global statistics directly from handwritten documents (pages) \cite{Brink2012Writer,Hannad2016Writer,He2014Delta,Newell2014,Nicolaou2015Sparse}. For example, the angles of stroke directions, the width of the ink trace, and the histograms of local binary patterns (LBP) and local ternary patterns (LTP) have been used for writer identification purposes. Allograph-based methods rely on local descriptors computed from small patches (allographs), and then a global document descriptor is statistically calculated using the local descriptors of one document \cite{Christlein2017Writer,Christlein2015Offline,He2015Junction}. These two methods can be further combined to form a discriminative global feature \cite{Bulacu2007Text,He2017Writer,Wu2014Offline}. The semi-supervised feature learning pipeline proposed in this work is based on allographs for offline writer identification.

Although writer identification has achieved excellent performance on some benchmark datasets, there are considerable challenges in real-world applications. First, the use of different pens, the physical condition of the writer, the presence of distractions (such as multitasking and noise), and the changes in writing style with age are key factors resulting in the unsatisfactory performance of writer identification. Second, the writers of the training set are different than those of the test set, and every writer only contributes a few handwritten text images in the typically used benchmark datasets. Third, the number of handwritten documents in benchmark datasets is highly insufficient for convolutional neural network (CNN) model training; therefore, training a reliable CNN model using limited data  is a challenge. Moreover, almost all published methods are based on supervised learning, which cannot achieve landmark results due to the limited amount of labeled data present in the benchmarks. Some researchers utilize different data augmentation methods to address these problems. However, these data augmentation methods that are used in writer identification easily lead to model overfitting and require a considerable amount of extra data. To overcome the aforementioned challenges and then tightly integrate with writer identification in practice, we propose a novel insight for writer identification.

CNNs are a well-known deep learning architecture inspired by the natural visual perception mechanism of living creatures. CNNs have been widely used and have achieved exciting performance in the fields of image classification, object recognition and object detection and tracking \cite{He2016Deep,Krizhevsky2012ImageNet,Simonyan2015Very,Szegedy2015Going} due to their powerful ability to learn deep features. The recent progress in writer identification is mainly attributed to advancements in CNNs based on supervised \cite{Christlein2014Writer,Christlein2017Writer,Christlein2015Offline,Fiel2015Writer,He2017Writer,Tang2017Text,Xing2016DeepWriter} and unsupervised feature learning \cite{Christlein2017Unsupervised}. The features extracted from CNNs perform better as discriminative characteristics compared to handcrafted features. For example, Xing and Qiao et al. \cite{Xing2016DeepWriter} designed a multistream CNN structure for writer identification and achieved a high identification accuracy on the IAM \cite{Marti2002The} and HWDB \cite{Liu2013Online} datasets using a small amount of handwritten documents. In \cite{Christlein2015Offline}, Christlein proposed using activation features from CNNs as local descriptors for writer identification and improved the identification performance on the ICDAR2013 dataset. R. Eldan et al. \cite{Eldan2015The} showed that a deeper network would learn a more discriminative representation but will need more resources to train. Therefore, we recommend that a tradeoff and a deep residual neural network with 50 layers (ResNet-50) could be applied in our work.

In contrast to the supervised learning approaches, semi-supervised learning significantly surpasses supervised learning when annotated data are limited in the training set, e.g., weakly labeled or unlabeled data \cite{Huang2017Semi,Weston2012Deep,Zhu2002Learning}.  In particular, semi-supervised learning saves the time and budget needed for annotating data when the volume of clean labeled data is limited. Some recent studies investigated a semi-supervised learning pipeline by combining unsupervised learning with supervised learning \cite{Rasmus2015Semi,Varior2016A} to assign an original label or a new label to unlabeled data \cite{Lee2013Pseudo,Odena2016Semi,Papandreou2016Weakly}.  Motivated by the previous studies, we attempt to use a modified semi-supervised learning method by assigning a weighted uniform label distribution to extra unlabeled data (extra data) according to the original labeled data (real data). We believe that the proposed approach has the potential to regularize the baseline for improving identification performance.

Therefore, we proposed a semi-supervised method that leverages a deep CNN and the weighted label smoothing regularization (WLSR) to form a powerful model that learns discriminative representations for offline writer identification in our work. Specifically, we first preprocess the original labeled data and the extra unlabeled data. Then, these original labeled data and extra unlabeled data are fed into a deep residual neural network (ResNet) \cite{He2016Deep} simultaneously. Furthermore, the WLSR method regularizes the learning process by integrating the unlabeled data, which can reduce the risk of overfitting and direct the model to learn more effective and discriminative features. Finally, the local features of every test handwritten document are extracted and encoded as a global feature vector for identification.  

To summarize, this study makes the following contributions:

\textbf{A}.	This study is a pioneering work that uses a semi-supervised feature learning pipeline to integrate extra unlabeled images and original labeled images into the ResNet model for writer identification. 

\textbf{B}.	The WLSR method of semi-supervised learning is used to regularize the identification model with unlabeled data. We thoroughly evaluate its availability on public datasets.

\textbf{C}.	Our results show that the proposed semi-supervised learning model had a consistent improvement over the deep residual neural network baseline and achieved better performance than existing approaches on benchmark datasets.

The remainder of this paper is organized as follows. Sec. 2 provides an overview of the related works in the field of writer identification. The process of the semi-supervised learning pipeline is presented in Sec. 3. The performance and evaluation are given in Sec. 4. Sec. 5 presents the discussion. Sec. 6 provides a summary and the outlook for future research.

\section{Related Work}
In this section, we review related work on writer identification that considered different data augmentation approaches to address cutting-edge challenges. Some researchers considered data augmentation in intrasets \cite{Christlein2015Offline,Fiel2015Writer,Tang2017Text,Xing2016DeepWriter}, but this easily led to model overfitting. Two recent studies added extra labeled data into the original data to enlarge the training set, which in turn required a vast amount of extra data to improve the identification results \cite{Christlein2014Writer,Christlein2017Writer}. 

S. Fiel et al. \cite{Fiel2015Writer} used a series of image preprocessing methods (binarization, text line segmentation, and sliding window) and then generated a discriminative feature by CaffeNet for each $56\times56$ image patch. Because CNNs have to be trained on a large amount of data to achieve a good result, he cut the line images into patches using a sliding window model with a step size of 20 pixels and rotated each patch of the sliding window from $-25$ to $+25$ degrees using a step size of 5 degrees. Thus, the new training set consists of more than 2,300,000 image patches, which artificially enlarged the original training set. His proposed algorithm achieved good performance on the ICDAR2011 \cite{Louloudis2011ICDAR} and CVL \cite{Kleber2013CVL} datasets, but this algorithm failed to improve the performance on the ICDAR2013 \cite{Louloudis2013ICDAR} dataset.
Furthermore, the CNN was trained on word images of the IAM dataset and the features of the CVL dataset extracted from the pretrained CNN. It suggested that the IAM and CVL datasets share a similar sample space. In \cite{Tang2017Text}, Tang introduced a new method for offline writer identification using a CNN and a joint Bayesian approach to contend with insufficient benchmark datasets for CNN model training. Tang also used words segmented from handwritten documents as elements to permute the texts to generate a significant number of images, which were subsequently converted to form handwritten pages. In addition, all the reconstructed handwritten pages were split into some nonoverlapping patches for training. In \cite{Xing2016DeepWriter}, Xing introduced a data augmentation method to enhance the performance of the proposed DeepWriter. However, these data augmentation methods only enlarged the dataset in the area of the intraset, and existing models did not consider dealing with the generated data, leading to an overfitting situation and limitations of feature learning in CNNs.

In \cite{Christlein2014Writer}, Christlein created a combined dataset (MERGED) consisting of 559 scribes with four documents per writer, resulting in 2236 documents from the ICDAR2013 and CVL datasets. Thereby, the training set was enlarged, and the outcomes on the MERGED datasets slightly differ from the image vocabularies that can be calculated from the ICDAR2013 experimental set or the CVL dataset. Furthermore, Christlein et al. \cite{Christlein2017Writer} showed that the identification rate on the CVL test set could be improved by adding additional datasets (ICDAR2011 and IAM \cite{Marti2002The}) into the CVL training set. Although existing data augmentation approaches have the capability to improve the identification performance using the extra data, we can imagine that it requires a large amount of extra labeled data. In practice, however, we do not have access to collect a large number of samples for writer identification.

In contrast to the aforementioned works, we employed a semi-supervised feature learning pipeline that allows adding data without a label. We assumed that the semi-supervised feature learning approach could effectively avoid overfitting and require less extra data to improve the ability of feature learning of the baseline.

\begin{figure*}[t]
    \centering    
    \includegraphics[scale=1.2,width=12.77cm,height=8.0cm]{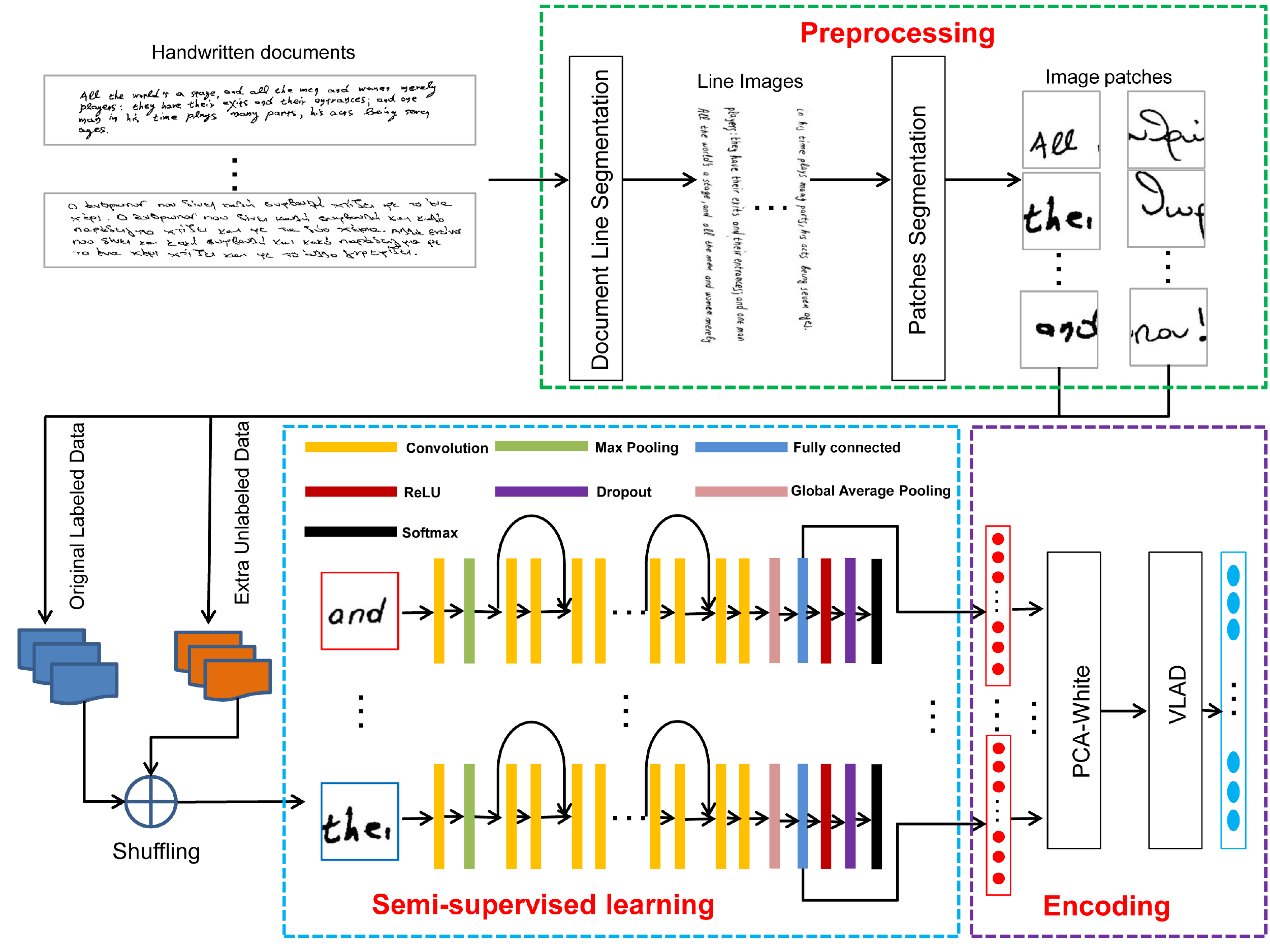}
    \caption{The pipeline of semi-supervised feature learning, which consists of three parts: preprocessing (green dotted box), semi-supervised learning (blue dotted box) and encoding (purple dotted box). During training, the original labeled data and extra unlabeled data are shuffled and fed into the semi-supervised learning network for training. For testing, the local features (red rectangles with solid edge in encoding part) of testing handwritten documents are extracted from the fully connected layer of the pretrained model, and then all the local features of one handwritten test document are encoded into a global feature vector (blue rectangles with solid edge in encoding part).}\label{Fig1}
\end{figure*}

\section{Semi-supervised Feature Learning Pipeline}

As shown in Fig. \ref{Fig1}, our proposed semi-supervised feature learning pipeline consists of three parts. \textbf{A}. Preprocessing: For the ICDAR2013 dataset, the handwritten documents are segmented into line images by a line segmentation method \cite{Srinivasan2007A}, and then the line images are split up using a sliding window approach without overlapping. For the IAM and CVL datasets, we normalize the word images already provided. \textbf{B}. Semi-supervised learning: During training, the original labeled data (real data) and extra unlabeled data (extra data) are shuffled, and then they are simultaneously fed into ResNet-50 baseline, which is regularized by WLSR. Furthermore, the trained model is used for extracting local features of testing handwritten documents. Specifically, all local features of handwritten test documents are extracted from the fully connected layer, and thus, all layers after the fully connected layer can be discarded. \textbf{C}. Encoding: We reduce the dimensions of local features with PCA-White \cite{J2012Negative}, and then the vector of locally aggregated descriptors (VLAD) \cite{Jegou2012Aggregating} is used to encode the local features of every test document as a global feature vector, which is used for writer identification with the nearest neighbor approach. All of the parts will be concretely introduced in the following.

\begin{figure*}[t]
    \centering    
    \includegraphics[scale=0.7,width=12.77cm,height=6.44cm]{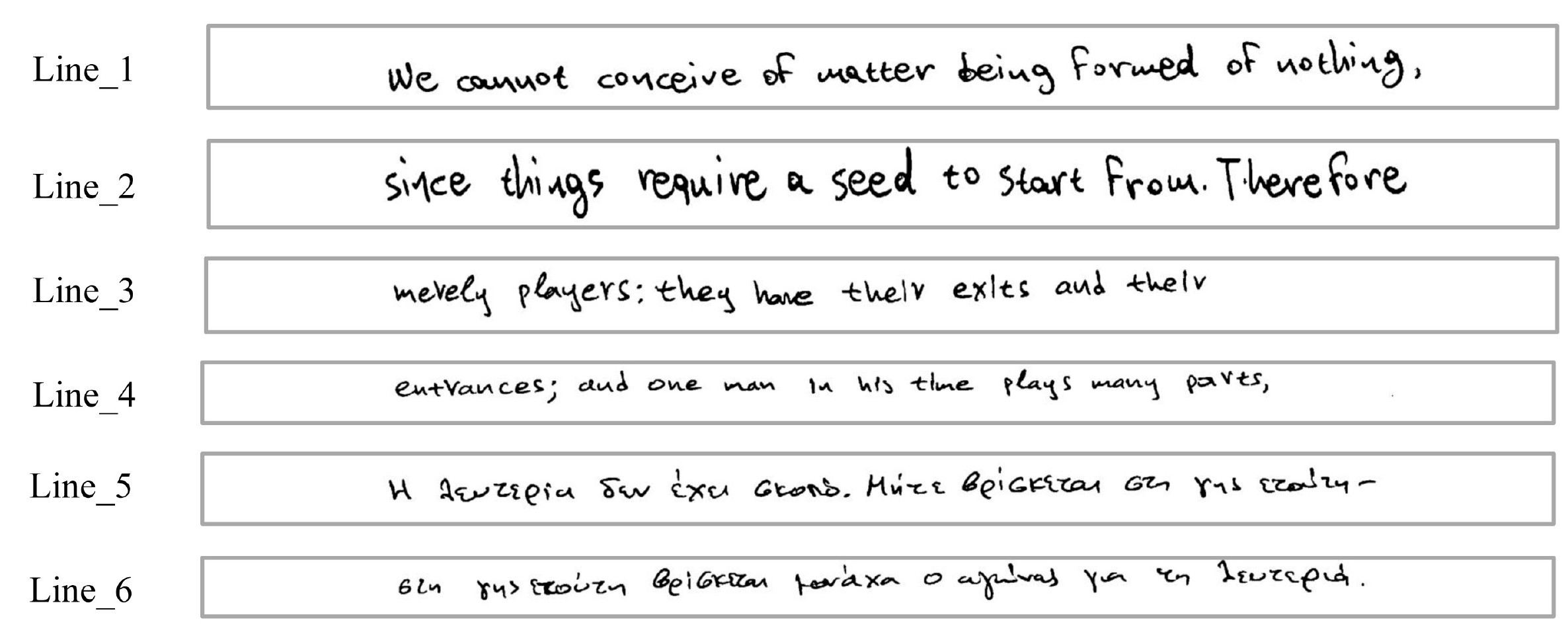}
    \caption{Part of the line images of the ICDAR2013 dataset are segmented by the proposed line segmentation approach and are normalized with their original aspect ratio.}\label{Fig2}
\end{figure*}
\begin{figure*}[t]
    \centering    
    \includegraphics[scale=0.7,width=12.77cm,height=8.44cm]{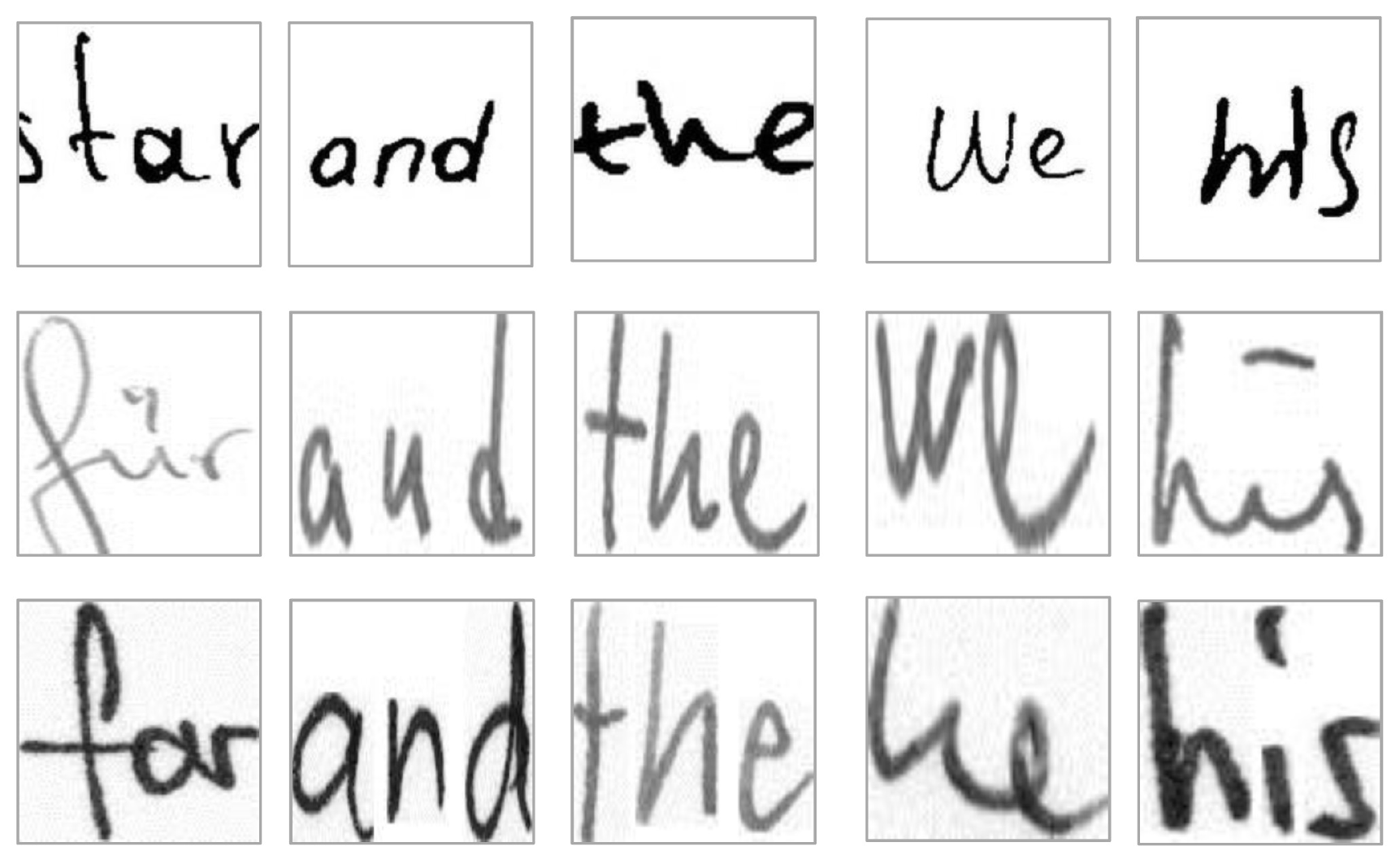}
    \caption{Part of the patches extracted from the ICDAR2013 dataset (top row), word images provided by the CVL dataset (middle row) and word images provided by the IAM dataset (bottom row), where all have been preprocessed. The patches of the ICDAR2013 dataset are normalized to $256\times256$. Each word image with size $x\times y$ in the CVL and IAM datasets is normalized to an image of size $256\times m$ or $m\times256$ such that $\frac{x}{y}=\frac{256}{m}$ or $\frac{x}{y}=\frac{m}{256}$.}\label{Fig3}
\end{figure*}

\subsection{Preprocessing}

First, a binarization is implemented for all handwritten pages with the Otsu \cite{Otsu2007A} method. Second, the handwritten pages have to be segmented. Because the CVL dataset \cite{Kleber2013CVL} and IAM \cite{Marti2002The} dataset already provide a segmentation of the words, these images are directly used for training and evaluating after normalization, as shown in Fig. \ref{Fig3}. For the ICADR2013 competition on the Writer Identification dataset \cite{Louloudis2013ICDAR}, the handwritten documents are segmented into lines with the method proposed by Arivazhagan \cite{Srinivasan2007A}. The line segmentation method is based on a statistical approach that segments the text lines exactly. In addition, we normalize the line images to a height of 256 pixels and maintain their aspect ratio. Finally, all text lines are cut into patches with a size of $256\times256$ without overlap using the sliding window approach. Some line images and patches of the ICDAR2013 dataset are shown in Fig. \ref{Fig2} and Fig. \ref{Fig3}, respectively. Furthermore, we remove noise patches (e.g., blank patches) to avoid adverse effects.

\subsection{Semi-supervised Learning}

In this section, we thoroughly introduce the process of the proposed semi-supervised learning. Semi-supervised learning is based on a baseline (ResNet-50) and WLSR method. The baseline serves as an identification model, and the local features of testing handwritten pages are extracted from the fully connected layer of the baseline during testing. WLSR regularizes the baseline and directs the model to learn more discriminative features. 

\subsubsection{CNN baseline}
K. He et al. \cite{He2016Deep} first proposed ResNet for image classification and object recognition and achieved exciting results, and then ResNet became widely used in other tasks due to its strong feature learning ability. In this work, ResNet-50 is used as a baseline because it learns discriminative representations without consuming too much of the time and computational budgets in writer identification. A ResNet consists of residual units that have two branches. One branch has several convolutional layers and learns the features of the input, and the other bypasses the other branch and forwards the result of the previous layer. These units help the CNN model preserve the identity and maintain a deeper structure. Following the conventional fine-tuning strategy, we use a model pretrained on ImageNet. To avoid model overfitting and to learn more discriminative features, we add a rectified linear unit (ReLU) layer \cite{Glorot2011Deep} and replace the original pooling layer with a global average pooling layer \cite{Lin2013Network} before the fully connected layer. In addition, we modify the last layer to have K neurons to predict the K classes, where K is the number of classes in the original training data. The extra data are mixed with the original data as the input of the CNN. That is, the original labeled training data and the extra unlabeled data are shuffled and simultaneously trained. After training, the local features of all test handwritten documents are extracted from the fully connected layer. Additional implementation details are provided in section 4.3.

\subsubsection{Weighted Label Smoothing Regularization Method}
Label smoothing regularization (LSR) was first used for fully supervised learning in the 1980s and was recently proposed to regularize the classifier layer by estimating the marginalized effect of label dropout during training \cite{Szegedy2016Rethinking}. In the person reidentification task, Zheng \cite{Zheng2017Unlabeled} extended LSR to label smoothing regularization for outliers (LSRO), which leveraged unsupervised data generated by GAN and set the virtual label distribution to be uniform over all classes, effectively regularizing the baseline model and achieving better retrieval performance than the baseline. In this work, we propose the WLSR method to regularize the CNN baseline with the extra unlabeled data for offline writer identification. WLSR sets the virtual label distribution to be a weighted uniform distribution over all classes, which effectively regularizes the baseline according to the original training data distribution. For instance, if the original training set has a large number of common features that do not benefit writer identification (e.g., some ink traces and scribe width), the identification model may be misdirected to take these common features as a discriminative representation, which limits the discriminative ability of the model. However, if we add these common features of extra unlabeled data into the model for training, the classifier will make an incorrect prediction toward the labeled words, and thus, the classifier will be penalized. Moreover, the regularization ability of WLSR is decided by the similarity of the sample space between the original labeled data and the extra unlabeled data. If the extra unlabeled data are located nearer the original training data in the sample space, the regularization ability of WLSR will be more effective. Otherwise, the performance of WLSR will be undesirable.

WLSR is proposed to be used with cross-entropy loss. Formally, let $k\in\{1,2,...,K\}$ be the original training data class and $N$ be the numbers of the original training data. The cross-entropy loss is shown in Eq. (\ref{Eq1}).
\begin{gather}
\label{Eq1}
\centering
l=-\sum\limits_{k=1}^{K}log(p(k))q(k),
\end{gather}
where $p(k)\in[0,1]$ is the predicted probability of training data belonging to class $k$, which is derived from the softmax function that normalizes the output of the previous CNN layer, and $q(k)$ is the ground-truth distribution. Let $y$ be the ground-truth class label. A pair $(x_i,y_i) $ is called the original training example, and $i\in \{1,2,...,N \}$.

\begin{figure*}[t]
	\subfigure[Label distribution of real data]{
		\begin{minipage}[t]{0.5\linewidth}
			\centering
			\includegraphics[width=2.3in]{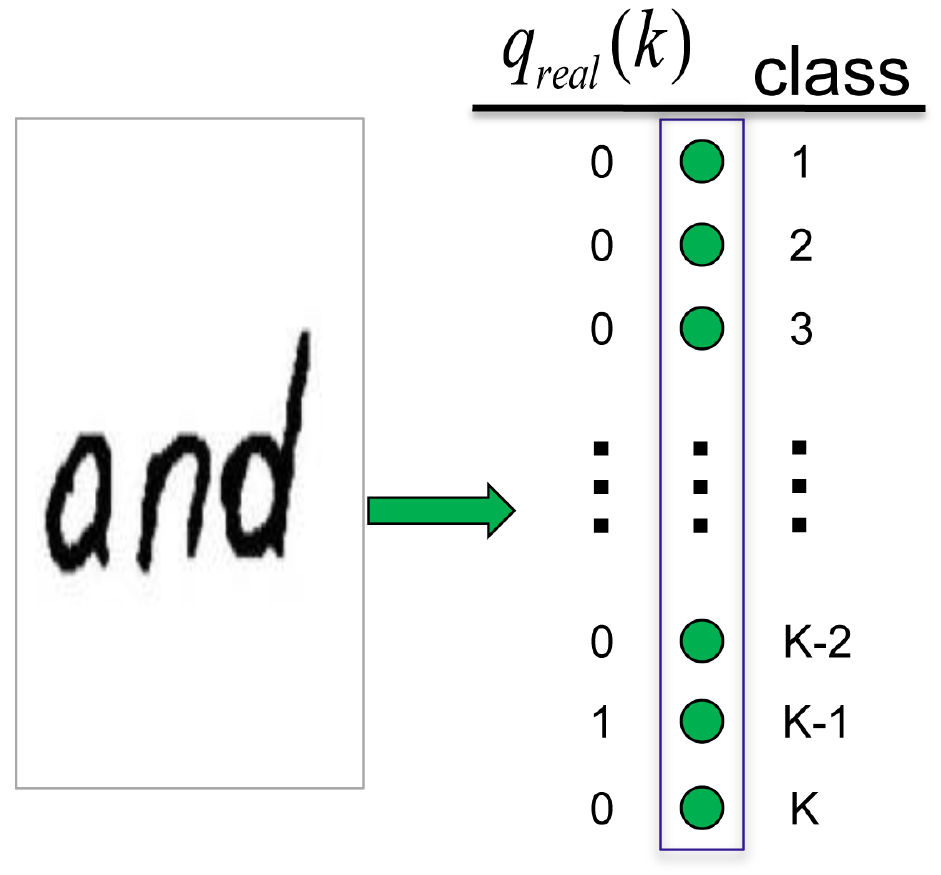}
			\label{Fig4_a} 
		\end{minipage}%
	}
	\subfigure[Label distribution of extra data]{
		\begin{minipage}[t]{0.5\linewidth}
			\centering
			\includegraphics[width=2.3in]{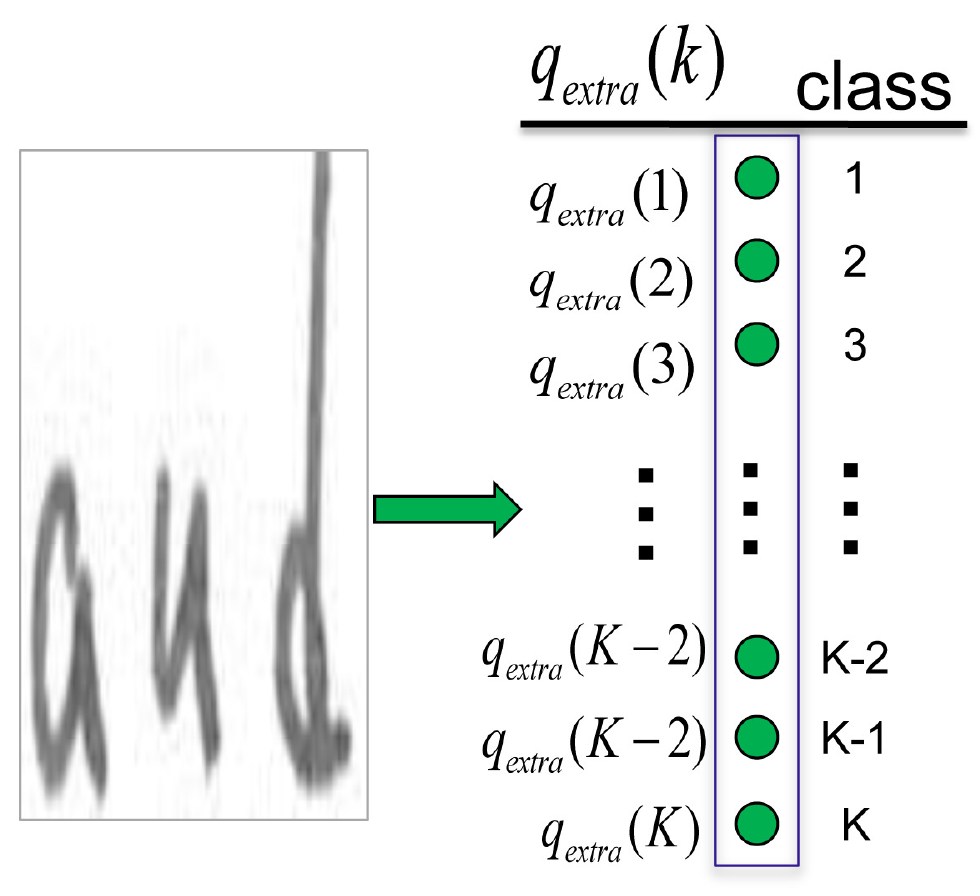}
			\label{Fig4_b} 
		\end{minipage}
	}
	\caption{The label distributions of real data and extra data used in our proposed semi-supervised feature learning pipeline. The cross-entropy loss combines them and will be simultaneously optimized (Eq. (\ref{Eq8})). (a) The label distribution of real data (Eq. (\ref{Eq2})) is a one-hot distribution, which shows that the original cross-entropy loss only takes the ground-truth term into account (Eq. (\ref{Eq3})). (b) We propose the virtual weighted uniform label distribution for the extra data (Eq. (\ref{Eq6})), which is assumed to not belong to any predefined training classes. All extra data will result in an incorrect prediction, and thus, the network will be penalized.}
	\centering
\end{figure*}

For the original labeled data of the training set, its ground-truth distribution $q_{real}(k)$ is shown in Fig. \ref{Fig4_a}. It can be formulated as:
\begin{gather}
\label{Eq2}
q_{real}(k)=\begin{cases}
0, & k\ne y;\\
1, & k=y.
\end{cases}
\end{gather}
Combining Eq. (\ref{Eq1}) and Eq. (\ref{Eq2}), the cross-entropy loss of real data $loss_{real}$ can be rewritten as:
\begin{gather}
\label{Eq3}
loss_{real}=-log(p(y)).
\end{gather}
From Eq. (\ref{Eq3}), it is clear that minimizing $loss_{real}$ is equivalent to maximizing the predicted probability of the ground-truth class.

However, LSR was proposed to take the distribution of non-ground-truth classes into consideration \cite{Szegedy2016Rethinking}. LSR discouraged the network from being confident toward its prediction. Formally, its label distribution $q_{LSR}(k)$ is formulated as:
\begin{gather}
\label{Eq4}
q_{LSR}(k)=\begin{cases}
\frac{\varepsilon}{K}, & k\ne y;\\
1-\varepsilon+\frac{\varepsilon}{K}, & k=y.
\end{cases}
\end{gather}
where $\varepsilon\in[0,1]$ is a smoothing parameter. Intuitively, if $\varepsilon$ is too large, the network may fail to predict the ground-truth label. Considering Eq. (\ref{Eq1}) and Eq. (\ref{Eq4}), the cross-entropy loss is written as:
\begin{gather}
\label{Eq5}
loss_{LSR}=-(1-\varepsilon)log(p(y))-\frac{\varepsilon}{K}\sum\limits_{k=1}^{K}log(p(k)).
\end{gather}
Thus, $loss_{LSR}$ not only takes the ground-truth class into account but also pays attention to other classes, which effectively avoids network overfitting.

We extend LSR from the supervised domain to the semi-supervised domain and propose weighted label smoothing (WLSR) to train the extra unlabeled data. Specifically, we set the virtual label distribution as a weighted uniform distribution over all classes for the extra unlabeled data according to the real data distribution, as shown in Fig. \ref{Fig4_b}. Thus, the label distribution of the extra data $q_{WLSR}(k)$ can be formulated as:
\begin{gather}
\label{Eq6}
q_{WLSR}(k)=\frac{\sum\limits_{n=1}^{N}I(y_n =k)}{N}.
\end{gather}
Thus, combining Eq. (\ref{Eq1}) and Eq. (\ref{Eq6}), the cross-entropy loss of extra data $loss_{extra}$ can be written as:
\begin{gather}
\label{Eq7}
loss_{extra}=-\sum_{k=1}^{K}log(p(k))\frac{\sum\limits_{n=1}^{N}I(y_n=k)}{N},
\end{gather}
where $I(y_n=k)$ is an indicator function. The proposed semi-supervised feature learning pipeline shuffles and simultaneously trains the real data and the extra data. Combining Eq. (\ref{Eq3}) and Eq. (\ref{Eq7}), we  rewrite the cross-entropy loss of semi-supervised feature learning $loss_{WLSR}$ as:
\begin{equation}
\begin{split}
\label{Eq8}
loss_{WLSR} &=-(1-Z)\cdot loss_{real}-Z \cdot loss_{extra}\\
&=-(1-Z) \cdot log(p(y))-Z \cdot \sum\limits_{k=1}^{K}log(p(k))\frac{\sum\limits_{n=1}^{N}I(y_n=k)}{N},
\end{split}
\end{equation}
where $Z$ is an indicator. For the extra data, $Z=1$. For the original training data, $Z=0$. Therefore, the proposed semi-supervised feature learning method has two types of losses: one for real images, and the other one for extra images.

\begin{figure*}[t]
	\centering	
	\includegraphics[scale=0.7,width=12.77cm,height=9.0cm]{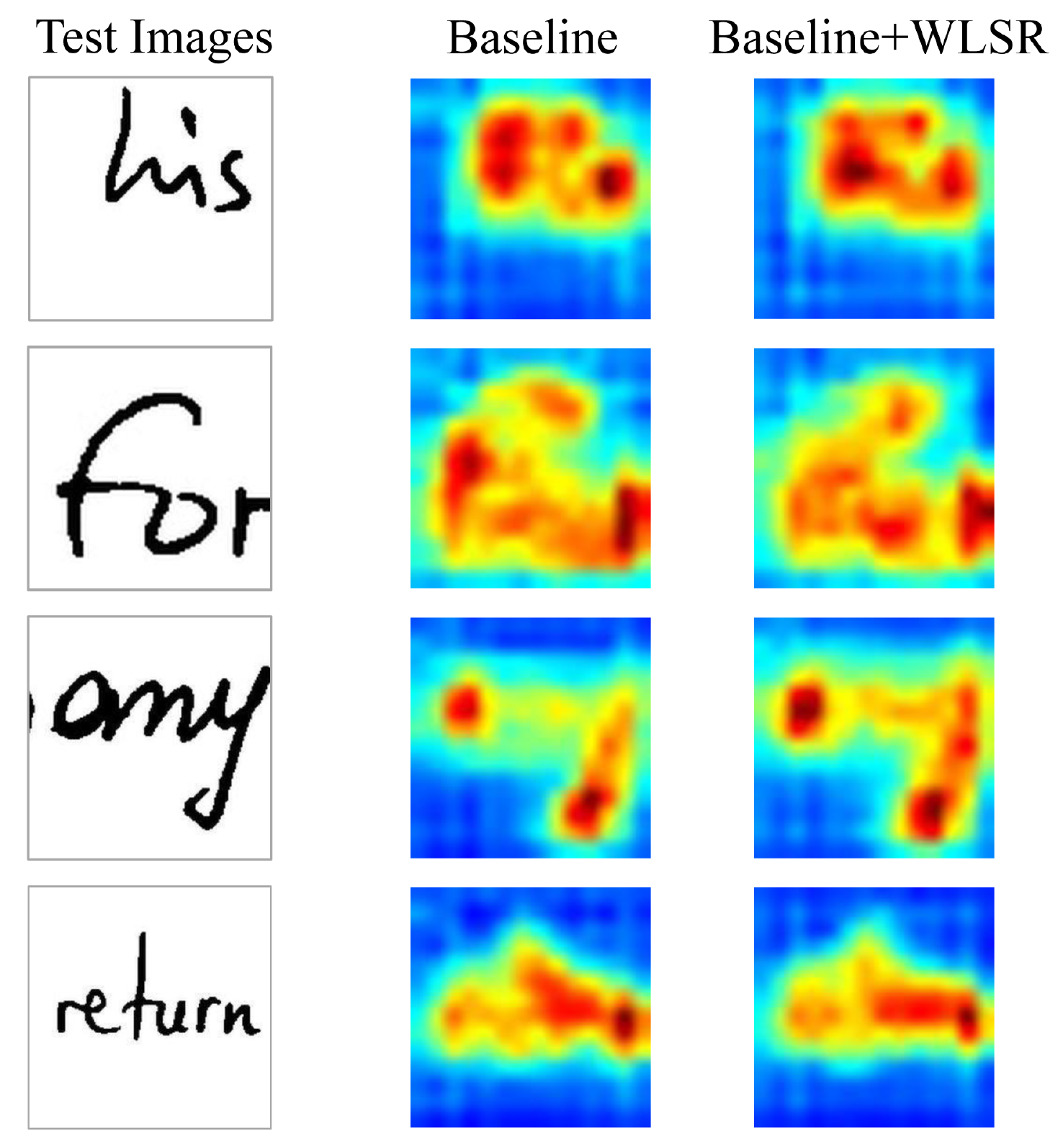}
	\caption{Visualization of the activation maps of the test patches of the ICDAR2013 test set in the baseline (ResNet-50) and the proposed semi-supervised learning model (baseline + WLSR). The baseline and the proposed semi-supervised learning network activate different patterns to the content of the patches. We can observe that the activation maps of the semi-supervised learning network more correctly and clearly show the contents of the test patches than the activation maps extracted from the baseline.}\label{Fig5}
\end{figure*}

To find the differences between the baseline ResNet-50 and the proposed semi-supervised learning pipeline baseline+WLSR, we visualize the intermediate feature maps of the two pretrained models. We take some patches of the ICDAR2013 test set for testing. The selected patches belong to various handwritten documents that perform poorly in the baseline, while they achieve the desired results in the semi-supervised learning model. For each patch, its activation is obtained from the intermediate layer ``res4fx'' of the network, the size of which is 14 * 14. Then, we visualize the sum of several activation maps. As shown in Fig. \ref{Fig5}, we observe that the baseline network and the proposed semi-supervised learning network activate different patterns in the content of patches. In particular, the activation maps of the semi-supervised learning more correctly and clearly exhibit the contents of test patches than the activation maps extracted from the baseline. That is, the representations of the semi-supervised learning model are more discriminative, which is why the proposed semi-supervised learning can produce better results than the baseline. 

\subsection{Encoding}

The all-local descriptors were extracted from the pretrained model during testing. We need to aggregate them to encode a global feature vector for each test document. First, we reduce the dimensionality of the local descriptors with PCA-White, which has been shown to effectively reduce the identification time and improve the identification performance \cite{Christlein2017Writer,Christlein2017Unsupervised}. In addition, we encode the all-local descriptors of each test page as the global feature vector with VLAD, which encodes the first-order statistics by aggregating the residuals of local features to their corresponding nearest cluster centroid. VLAD is a standard encoding method and has been widely used in writer identification \cite{Christlein2015Writer,Christlein2017Unsupervised} and other information retrieval tasks \cite{Chattopadhyay2016Supervised,Paulin2016Convolutional}. Formally, a codebook $D=\{c_1,c_2,...,c_k\}$ is first computed by k-means with $k$ centroids, and all $S$ local features $f_S\in R^m$ of every test handwritten image are assigned to their nearest cluster centroid. Then, all residuals between the cluster centroid and the assigned local features are accumulated for each cluster:
\begin{gather}
\label{Eq9}
v_k=\sum\limits_{f_S:NN(f_S)=c_k}^{}(f_S-c_k),
\end{gather}
where $NN(f_S)$ refers to the nearest neighbor of $f_S$ in dictionary $D$. All $v_k$ are concatenated as a global feature vector of one handwritten page:
\begin{gather}
\label{Eq10}
v=(v_1^T,v_2^T,...,v_K^T)^T.
\end{gather}
Thus, the global feature of each test document will eventually be $km$-dimensional.

\section{Evaluation}

In the following sections, we describe the datasets and evaluation metrics that we used for evaluating our proposed method. Then, we verify that WLSR has the potential to regularize the baseline for improving identification performance. Furthermore, we show the impacts of using various dimensions of local features, different numbers of extra unlabeled data during training and different centroids of k-means during encoding. Finally, we compare our method to other methods for writer identification.

\subsection{Datasets}
There are three different benchmark datasets used for evaluation: the ICDAR2013  dataset\footnote{http://rrc.cvc.uab.es/} \cite{Louloudis2013ICDAR}, the CVL dataset\footnote{https://cvl.tuwien.ac.at/research/cvl-databases/} \cite{Kleber2013CVL} and the IAM dataset\footnote{http://www.fki.inf.unibe.ch/databases/iam-handwriting-database/} \cite{Marti2002The}. All of these datasets are public and have been used in many recent publications \cite{Christlein2014Writer,Christlein2017Writer,Fiel2015Writer,Nicolaou2015Sparse,Tang2017Text,Xing2016DeepWriter}. Of note, Fiel \cite{Fiel2015Writer} trained the network on the IAM dataset and evaluated on the CVL dataset, achieving good performance. The results suggested that the word images in the IAM and CVL datasets can share a more similar sample space. Tang \cite{Tang2017Text} trained his model on the ICDAR2013 dataset, tested on CVL the dataset and provided an impressive identification effect, which revealed that the patches of the CVL and ICDAR2013 datasets have a highly similar sample space. Therefore, we take IAM word images and CVL patches as the extra unlabeled data to evaluate CVL word images and ICDARA2013 patches, respectively.

\textbf{ICDAR2013 \cite{Louloudis2013ICDAR}:} The ICDAR2013 benchmark dataset is divided into a training set with documents written by 100 writers and a test set with documents written by 250 writers. Every writer contributed four documents, including two Greek documents and two English documents.

\textbf{CVL \cite{Kleber2013CVL}:} There are 310 writers who contributed documents for the CVL dataset. The 27 writers of the training set contributed seven documents each, and the 283 writers of the test set contributed five documents each. All writers contributed one German document, and the others are English documents.

\textbf{IAM \cite{Marti2002The}:} The IAM dataset was contributed to by approximately 400 writers with 1066 forms. In the collection, 82,227-word examples are built from a vocabulary of 10,841 words. All of the documents were written in English.

\subsection{Evaluation Metrics}
The mean average precision (mAP) and hard TOP-k, which are common evaluation metrics in image and information retrieval tasks, are used for our experimental evaluation.

A ranked list of all documents in the query library is generated according to the similarity of each query document. Suppose that there are $N$ handwritten documents from the query; thus, the average precision $AP(i)$ of the $i_{th}$ ($1 \leq i \leq N$) query document is Eq. (\ref{Eq11}).
\begin{gather}
\label{Eq11}
AP(i)=\frac{\sum_{k=1}^{M}P(k)\cdot rel(k)}{R} 
\end{gather}
where $M$ is the number of documents in the query library and $R$ is the number of relevant documents of the $i_{th}$ query document in the query library. $P(k)$ is the precision at rank $k$, which is given by the number of documents from the same writer in the query up to rank $k$ divided by $k$. $rel(k)$ is an indicator function, where $rel(k)=1$ when the document retrieved at rank $k$ is from the same writers, and $rel(k)=0$ otherwise. 

The mAP is the mean value of the average precision of all query documents. It can be written as:
\begin{gather}
	mAP=\frac{\sum\limits_{i=1}^{N}AP(i)}{N}. 
\end{gather}
The hard TOP-k depends on the calculation of the percentage of the query result, where the $k$ highest ranked documents are from the same writer. 

\subsection{Experiments}

The proposed method was evaluated on the ICDAR2013, CVL and IAM benchmark datasets. We present the implementation details and analysis of the experimental results in the following.

\subsubsection{Implementation Details}
In this work, we adopt the ResNet-50 model as a baseline. To gather more abstract features, we take the global average pooling layer to replace the original pooling layer and add a ReLU activation feature layer. Furthermore, the last fully connected layer was modified to have 100 and 27 neurons for ICADAR2013 and CVL, respectively. We add a dropout layer before the last convolutional layer and set the dropout rate to 0.5 for training. The momentum of stochastic gradient descent is set to 0.9. We set the learning rate of the convolutional layers to 0.1 and have it decay to 0.01 after 45 epochs. To evaluate ICDAR2013, we take the ICDAR2013 training image patches as the original labeled data and the CVL training image patches as the extra unlabeled data. The CVL and IAM datasets already provide a segmentation of words. Thus, we directly take the CVL training words as the original labeled data and the IAM words as the extra unlabeled data to evaluate the CVL dataset. The size of the segmented image patches is set to $256\times256$, while the width or height of word images was set to 256 pixels and the original aspect ratio was maintained. We extracted the local features of the test images in the first fully connected layer. The similarity between two handwritten documents was calculated by the Euclidean distance for ranking.

\begin{figure}[t]
	\centering	
	\includegraphics[scale=1.0,height=6cm,width=10cm]{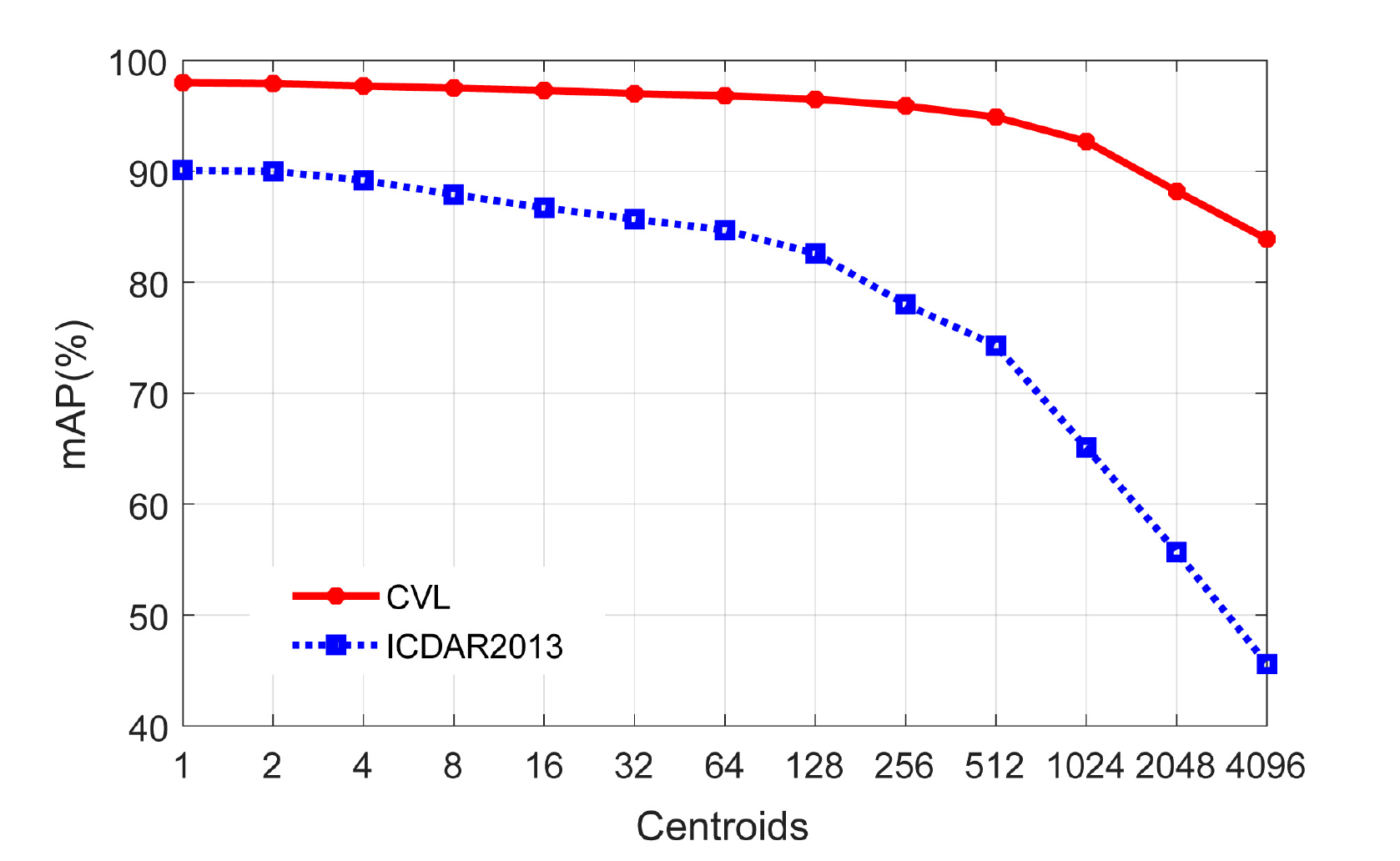}
	\caption{The influence of the number of centroids during encoding with VLAD. The mAP of the CVL dataset (red solid line) and the ICDAR2013 dataset (blue dotted line) exchange with the number of centroids of k-means.}\label{Fig6}
\end{figure}

\begin{table}[t]
	\centering
	\caption{The influence of the number of neurons of the fully connected layer on the CVL test set evaluated with the hard TOP-k and mAP metrics (\%).}\label{Table1}
	\setlength{\tabcolsep}{5mm}{
	\begin{tabular}{lccccc}
		\hline
		& TOP-1 & TOP-2 & TOP-3 & TOP-4 & mAP \\
		\hline
		Fc-512 & 97.9 & 97.0 & 93.6 & 85.0 & 96.4 \\
		
		Fc-1024 & 98.4 & 97.4 & 94.9 & 87.9 & 97.0 \\
		
		\textbf{Fc-2048} & \textbf{99.2} & \textbf{98.2} & \textbf{96.0} & \textbf{90.2} & \textbf{98.0}\\
		
		Fc-4096 & 98.5 & 97.6 & 94.7 & 88.0 & 97.3 \\
		\hline
	\end{tabular}}
\end{table}

\subsubsection{Experimental Results}
First, we evaluate how the number of neurons of the fully connected layer affects writer identification. The number of neurons is set to 512, 1024, 2048, and 4096, which are assessed on the CVL dataset, as shown in Table \ref{Table1}. It was evident that the semi-supervised feature learning pipeline achieves the best performance on hard TOP-k and mAP metrics when the number of neurons of the first fully connected layer is set to 2048. Thus, all the following experiments use this configuration. 

\begin{table}[t]
	\centering
	\caption{Comparison: the proposed semi-supervised feature learning vs. baseline on the CVL and ICDAR2013 test sets}\label{Table2}
	\setlength{\tabcolsep}{0.4mm}{
		\begin{tabular}{l|lccccc}
			\hline
			& & TOP-1 & TOP-2 & TOP-3 & TOP-4 & mAP \\
			\hline
			
			& 0 (baseline) & 98.3 & 97.0 & 92.5 & 87.0 & 95.7 \\
			
			\textbf{CVL}&12000 (baseline) & 98.4 & 97.0 & 94.0 & 87.2 & 96.8 \\

			&\textbf{12000(baseline+WLSR)} & \textbf{99.2} & \textbf{97.9} & \textbf{96.0} & \textbf{90.2} & \textbf{97.8} \\
			
			\hline
			
			& 0 (baseline) & 94.9 & 74.6 & 55.1 & N/A & 88.0 \\
			
			\textbf{ICDAR2013}&1000 (baseline) & 95.1 & 74.3 & 57.3 & N/A & 88.1 \\
			
			&\textbf{1000 (baseline+WLSR)} & \textbf{96.6} & \textbf{79.0} & \textbf{61.1} & N/A & \textbf{90.1} \\
			\hline
	\end{tabular}}
\end{table}

\begin{table}[t]
	\centering
	\caption{Comparison of the effect of various numbers of extra unlabeled images on the CVL test set evaluated with the hard TOP-k and mAP metrics (\%).}\label{Table3}
	\setlength{\tabcolsep}{2.41mm}{
	\begin{tabular}{lccccc}
		\hline
		& TOP-1 & TOP-2 & TOP-3 & TOP-4 & mAP \\
		\hline
		0 (baseline+WLSR) & 98.3 & 97.0 & 92.5 & 87.0 & 95.7 \\
		
		1000 (baseline+WLSR) & 98.8 & 97.9 & 95.0 & 88.5 & 97.3 \\
		
		5000 (baseline+WLSR) & 98.9 & 97.9 & 95.4 & 88.9 & 97.5 \\
		
		\textbf{12000 (baseline+WLSR)} & \textbf{99.2} & \textbf{97.9} & \textbf{96.0} & \textbf{90.2} & \textbf{97.8} \\
		
		24000 (baseline+WLSR) & 99.0 & 97.9 & 95.2 & 89.9 & 97.6 \\
		\hline
	\end{tabular}}
\end{table}

\begin{table}[t]
	\centering
	\caption{Comparison of the effects of the numbers of extra unlabeled images on the ICDAR2013 test set evaluated with the hard TOP-k and mAP metrics (\%).}\label{Table4}
	\setlength{\tabcolsep}{4.25mm}{
	\begin{tabular}{lcccc}
		\hline
		& TOP-1 & TOP-2 & TOP-3 & mAP \\
		\hline
		0 (baseline+WLSR) & 94.9 & 74.6 & 55.1 & 88.0 \\
		
		500 (baseline+WLSR) & 94.8 & 75.5 & 56.3 & 88.1  \\
		
		\textbf{1000 (baseline+WLSR)} & \textbf{96.6} & \textbf{79.0} & \textbf{61.1} & \textbf{90.1} \\
		
		2000 (baseline+WLSR) & 96.5 & 78.6 & 59.6 & 90.0 \\
		
		5000 (baseline+WLSR) & 94.9 & 74.3  & 56.5  & 88.0  \\
		\hline
	\end{tabular}}
\end{table}

\begin{table}[t]
	\centering
	\caption{Comparison of the performance with other methods on the CVL test set. Hard TOP-k and mAP metrics are listed (\%).}\label{Table5}
	\setlength{\tabcolsep}{3.5mm}{
		\begin{tabular}{lccccc}
			\hline
			& TOP-1 & TOP-2 & TOP-3 & TOP-4 & mAP \\
			\hline
			CS-UMD \cite{Kleber2013CVL} & 97.9 & 90.0 &71.2 & 48.3 & N/A \\
			
			QUQA A \cite{Kleber2013CVL} & 30.5 & 5.7 & 0.5 & 0.1 & N/A \\
			
			QUQA B \cite{Kleber2013CVL} & 92.9 & 84.9 & 71.5 & 50.6 & N/A \\
			
			TEBESSA-c \cite{Kleber2013CVL} & 97.6 & 94.3 & 88.2 & 73.9 & N/A\\
			
			TSINGHUA \cite{Kleber2013CVL} & 97.7 & 95.3 & 94.5 & 7.30 & N/A \\
			
			Fiel et al. \cite{Fiel2015Writer} & 98.9 & 97.6 & 93.3 & 79.9 & N/A \\
			
			Christlein et al. \cite{Christlein2014Writer} & 99.2 & 98.1 & 95.8 & 88.7 & 97.1 \\
			
			Nicolaou et al. \cite{Nicolaou2015Sparse} & 99.0 & 97.7 & 95.2 & 86.0 & N/A \\
			
			Christlein et al \cite{Christlein2017Writer} & 98.8 & 97.8 & 95.3 & 88.8 & 96.4 \\
			
			\textbf{Ours (single)} & 99.2 & 97.9 & 96.0 & 90.2 & 97.8 \\
			
			\textbf{Ours (2-streams)} & \textbf{99.2} & \textbf{98.4} & \textbf{96.1} & \textbf{91.5}\qquad & \textbf{98.0 }\\
			\hline
	\end{tabular}}
\end{table}

Second, we analyze the influence of the number of centroids $k$ during encoding with VLAD. In general, when k is larger, the retrieval performance is better for a large dataset. The experimental results on the ICDAR2013 and CVL datasets are shown in Fig. \ref{Fig6}. As shown, when the number of centroids is set to 1, we achieve the largest mAP (98.0\% and 90.1\% on ICDAR2013 and CVL, respectively). Moreover, the mAP of the two benchmarks consistently decreases as the number of centroids increases. Three reasons may explain the experimental results: \textbf{A}. The ICDAR2013 and CVL datasets are too small; therefore, they do not need more image vocabulary to represent themselves. \textbf{B}. Every writer wrote the documents with the same content in one dataset, which means that the diversity of the dataset is limited. \textbf{C}. The dimensions of the local feature are so large (2048 in this work compared to 64 in \cite{Jegou2012Aggregating}) that the local features are discriminative.

Third, we verify the regularization ability of the WLSR method in the semi-supervised feature learning pipeline. The same extra labeled and unlabeled data were added into the supervised baseline and the proposed semi-supervised pipeline for training, respectively. As shown in Table \ref{Table2}, the extra labeled data added in the baseline have almost no effect on writer identification, while the semi-supervised learning pipeline takes the same unlabeled data to improve the identification rate (on the CVL and ICDAR2013 datasets), which shows that the regularization of WLSR improves the performance of the baseline.

Moreover, we compare the proposed semi-supervised learning pipeline with the baseline. As shown in Table \ref{Table2},  when we add 12000 extra unlabeled IAM words into the CNN for training, our method significantly improves the writer identification performance on the CVL test set, which reveals that the WLSR method achieves improvements of 0.9\% (from 98.3\% to 99.2\%), 0.9\% (from 97.0\% to 97.9\%), 3.5\% (from 92.5\% to 96.0\%), 3.2\% (from 87.0\% to 90.2\%) and 2.1\% (from 95.7\% to 97.8\%) in hard TOP-1, hard TOP-2, hard TOP-3, hard TOP-4, and mAP, respectively. On ICADAR2013, we observe improvements of 1.7\%, 4.4\%, 6.0\% and 2.1\% in hard TOP-1, hard TOP-2, hard TOP-3, and mAP, respectively, when 1000 extra unlabeled CVL patches are added in ICDAR2013, as shown in Table \ref{Table2}. Thus, it is evident that the proposed semi-supervised feature learning pipeline effectively improves the performance of the baseline.

\begin{table}[t]
	\centering
	\caption{Comparison of the performance with the other methods on the ICDAR2013 test set. Hard TOP-k and mAP metrics are shown (\%).}\label{Table6}
	\setlength{\tabcolsep}{5.5mm}{
		\begin{tabular}{lcccc}
			\hline
			& TOP-1 & TOP-2 & TOP-3 & mAP \\
			\hline
			CS-UMD-b \cite{Louloudis2013ICDAR} & 95.0 & 20.2 & 8.4  & N/A \\
			
			HIT-ICG \cite{Louloudis2013ICDAR} & 94.8 & 63.2 & 36.5 & N/A \\
			
			TEBESSA-c \cite{Louloudis2013ICDAR} & 93.4 & 62.6 & 36.5 &  N/A\\
			
			CVL-IPK \cite{Louloudis2013ICDAR} & 90.9 & 44.8 & 24.5 &  N/A \\
			
			Fiel et al. \cite{Fiel2015Writer} & 88.5 & 40.5 & 15.8 & N/A \\
			
			Christlein et al. \cite{Christlein2014Writer} & 97.1 & 42.8 & 23.8  & 67.1 \\
			
			Nicolaou et al. \cite{Nicolaou2015Sparse} & 97.2 & 52.9 & 29.2 & N/A \\
			
			Christlein et al. \cite{Christlein2017Writer} & \textbf{98.2} & 71.2 & 47.7 & 81.4 \\
			
			\textbf{Ours (single)} & 96.6 & 79.0 & 61.1 & 90.1 \\
			
			\textbf{Ours (2-streams)} & 97.7 & \textbf{83.3}  & \textbf{63.7}  & \textbf{91.8} \\
			\hline
	\end{tabular}}
\end{table}

In addition, we find that the amount of extra unlabeled data profoundly affects the regularization ability of WLSR. If too little extra unlabeled data are incorporated into the pipeline, the regularization of the WLSR is insufficient. In contrast, if too much extra unlabeled data are added, the pipeline tends to assign weighted uniform prediction probabilities to all training data, as shown in Table \ref{Table3} and Table \ref{Table4}. Therefore, the appropriate amount of extra unlabeled data that should be added to the system varies by dataset to avoid poor regularization and pipeline overfitting.

\begin{figure*}[htbp] 
	\centering 
	\subfigure{ 
		\includegraphics[scale=0.7,width=12.82cm,height=8.4cm]{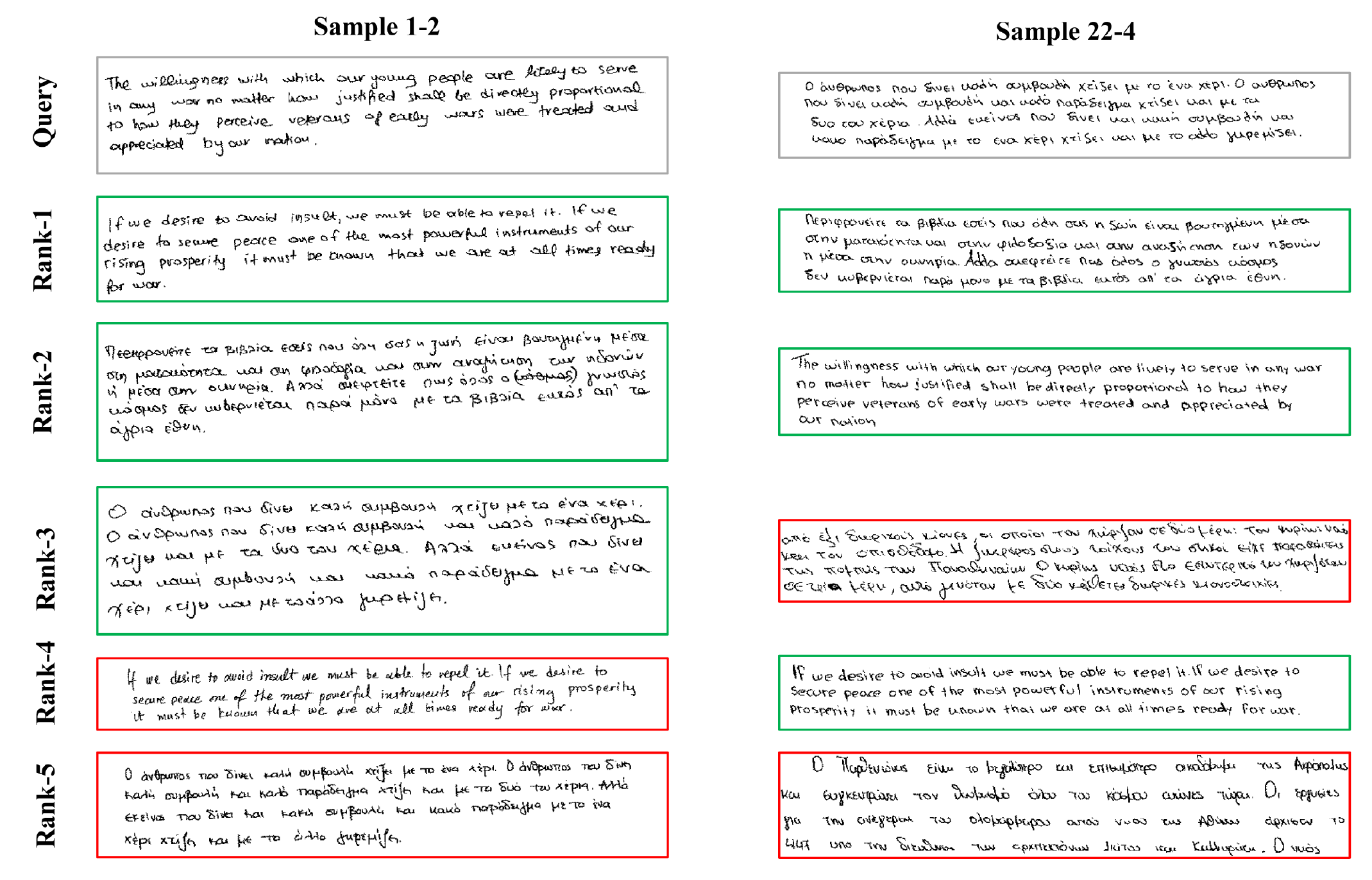} 
	} 
	\subfigure{ 
		\includegraphics[scale=0.7,width=12.56cm,height=8.4cm]{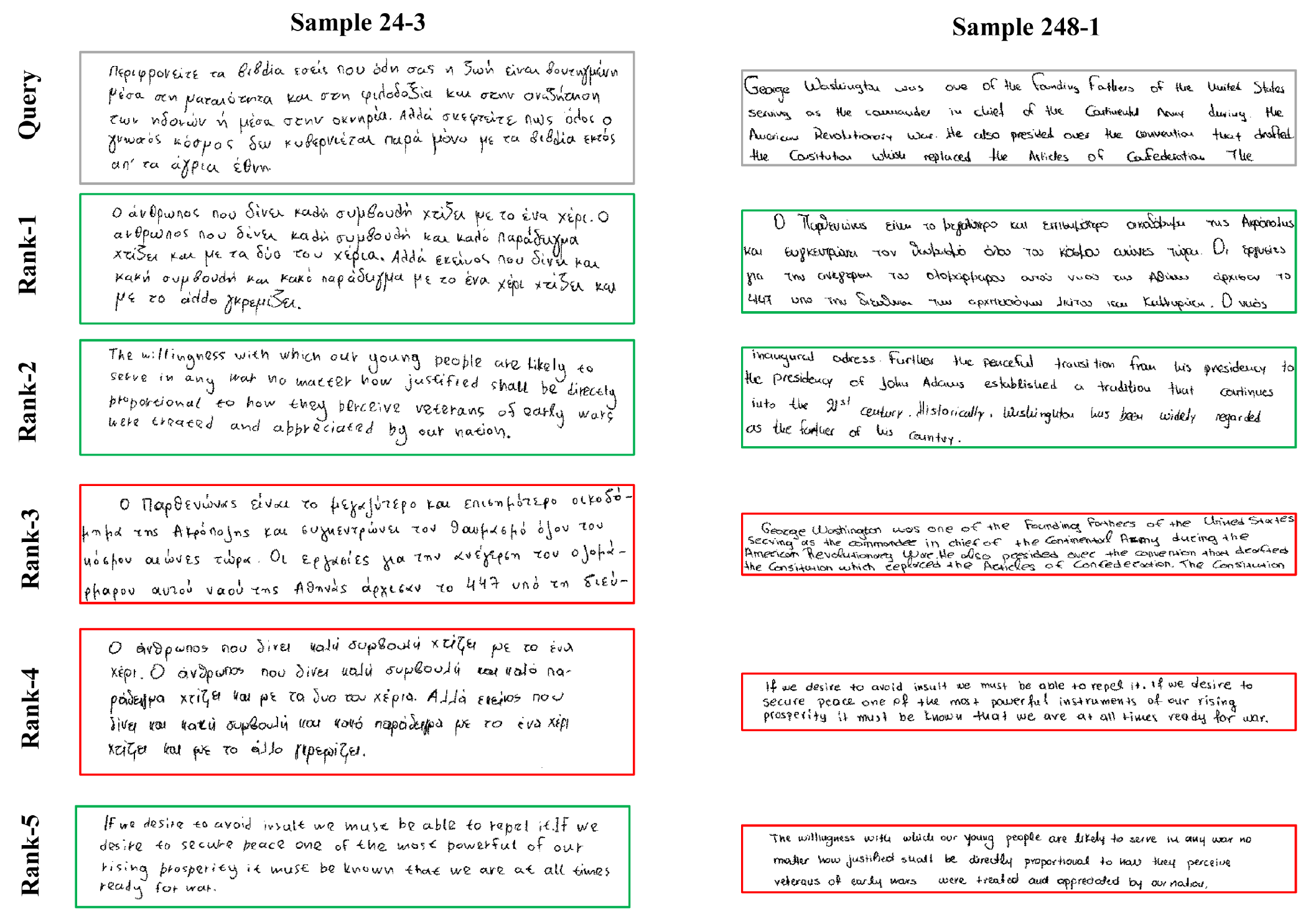} 
	}
	\caption{Writer identification results of the proposed semi-supervised feature learning method (single) on the ICDAR2013 dataset (sample 1-2, sample 22-4, sample 24-3, and sample 248-1). The images (gray border) are the query images. The identification images are sorted according to the similarity scores from top to bottom (from Rank-1 to Rank-5). We maintain the original aspect ratio of the images. } 
	\label{Fig7} 
\end{figure*}

Finally, we combined the two models generated by our method to form an ensemble (2-stream) to further enhance the identification performance and compared our proposed method with the other published methods on the ICDAR2013 and CVL datasets, as listed in Table \ref{Table5} and Table \ref{Table6}, respectively. We can observe that the semi-supervised learning pipeline can achieve a better result than most other supervised approaches. On the CVL dataset, we achieve hard TOP-1=99.2\%, hard TOP-2=98.4\%, hard TOP-3=96.1\%, hard TOP-4=91.5, and mAP=98.0\%, which are better results that those achieved by the other supervised methods. On ICDAR2013, we achieved hard TOP-1=97.7\%, hard TOP-2=83.3\%, hard TOP-3=63.0, and mAP=91.1\%, which are also very competitive results compared to the results of the other methods. In particular, the proposed semi-supervised learning method produces the desired performance on the ICDAR2013 test set with few extra unlabeled patches of the CVL training set, while Christlein et al. \cite{Christlein2014Writer} added the entire CVL training set into ICDAR2013 for training and achieved ordinary results. The results in Table \ref{Table5} and Table \ref{Table6} show that the semi-supervised feature learning method takes full advantage of the extra data, and it is more conveniently used in practice than other supervised methods \cite{Christlein2014Writer,Christlein2017Writer,Christlein2015Offline,Fiel2015Writer,Kleber2013CVL,Louloudis2013ICDAR,Nicolaou2015Sparse}. Fig. \ref{Fig7} presents some identification results achieved by the proposed semi-supervised feature learning method (single) on the ICDAR2013 dataset (sample 1-2, sample 22-4, sample 24-3, and sample 248-1). The images (gray border) are the query images. The identification images (red border and green border) are sorted according to the similarity scores from top to bottom (from Rank-1 to Rank-5). Images with a green border are correct candidates, and images with a red border images are incorrect candidates. Most ground-truth candidate images are correctly identified.

\section{Discussion}

In this study, we visualized the intermediate feature maps of the baseline and semi-supervised feature learning pipeline (Sec. 3.2.2). It showed that the activation maps of the semi-supervised learning more correctly show the contents of test patches than the activation maps extracted from the baseline. Then, we analyzed the impact of the dimensions of the local features, the centroids of VLAD encoding and the amount of extra unlabeled data (Sec. 4.3.2). Moreover, we experimentally showed that the proposed method could significantly improve the baseline and perform competitively with existing writer identification approaches, which benefit from the potential of regularization of WLSR. WLSR takes full advantage of extra unlabeled data for regularizing the baseline, and thus, the CNN learns effective and discriminative features. 

Due to some common representations in the extracted features, some researchers combined multiple handcrafted elements to derive a more reliable discriminative feature, yet restraining the impact of common features. For example, Helli extracted features using Gabor and XGabor filters and then developed a feature relation graph \cite{Helli2010A}. Considering the width of ink traces, a powerful source of information for offline writer identification consisted of a powerful feature (Quill) by combining with directions \cite{Brink2012Writer}. In \cite{He2015Junction}, they proposed a novel junction detection method for writer identification using stroke-length distribution and direction of ink of texts. Motivated by the above methods, we proposed a WLSR method to regularize and penalize the common features that are automatically learned features by the CNN and reducing the negative influence of common features.

To be honest, our proposed semi-supervised feature learning has a limitation in that WLSR depends on the similarity of the sample space between the original labeled data and extra unlabeled data. In the future, the generative adversarial networks (GANs), a system of two neural networks competing with each other in a zero-sum game framework, may be a potential choice to overcome this limitation. Because we can generate data by GANs and original data share the same sample space, we do not require any extra data from other datasets.

\section{Conclusion}
In this paper, we proposed a semi-supervised feature learning pipeline for offline writer identification. To the best of our knowledge, this is the first attempt to apply semi-supervised feature learning in the field of writer identification. Of note, the WLSR method is introduced to train the extra unlabeled data and the original labeled data simultaneously for the semi-supervised learning pipeline with regularization ability, which improved the identification results of the baseline model and achieved better performance than other popular methods on the CVL and ICDAR2013 datasets. 

In the future, we will choose a better encoding method that is suitable for small datasets of writer identification tasks to replace VLAD. Furthermore, we will adopt the unlabeled data generated by GANs to train the semi-supervised learning network because the generated data share a similar sample space with the original labeled data.

\section*{References}

\bibliographystyle{abbrv} 
\bibliography{reference}
\clearpage

\end{document}